\documentclass[sigconf]{acmart}

\usepackage{subcaption}
\usepackage{tabularx}
\usepackage{color, colortbl}
\usepackage{multirow}
\usepackage{booktabs}
\usepackage{ragged2e}
\usepackage{balance}

\definecolor{LightCyan}{rgb}{0.88,1,1}

\usepackage{scalerel,xparse}
\NewDocumentCommand\emojifrown{}{
    \scalerel*{
        \includegraphics{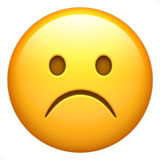}
    }{X}
}

\DeclareUnicodeCharacter{2639}{\emojifrown}
\DeclareUnicodeCharacter{FE0F}{}

\AtBeginDocument{%
  }

\copyrightyear{2024}
\acmYear{2024}
\setcopyright{rightsretained}
\acmConference[CIKM '24]{Proceedings of the 33rd ACM International
Conference on Information and Knowledge Management}{October 21--25,
2024}{Boise, ID, USA}
\acmBooktitle{Proceedings of the 33rd ACM International Conference on
Information and Knowledge Management (CIKM '24), October 21--25, 2024,
Boise, ID, USA}
\acmDOI{10.1145/3627673.3679635}
\acmISBN{979-8-4007-0436-9/24/10}
\begin{document}

\title{3M-Health: Multimodal Multi-Teacher Knowledge Distillation for Mental Health Detection}
\titlenote{\textcolor{red}{Warning: This paper contains examples that show suicide ideation and depression.}}


\author{Rina Carines Cabral}
\affiliation{%
    \institution{The University of Sydney}
    \city{Sydney}
    \state{NSW}
    \postcode{2006}
    \country{Australia}
}
\email{rcab5321@uni.sydney.edu.au}
\orcid{0000-0003-3076-0521}

\author{Siwen Luo}
\affiliation{
    \institution{The University of Western Australia}
    \city{Perth}
    \state{WA}
    \postcode{6009}
    \country{Australia}
}
\email{siwen.luo@uwa.edu.au}
\orcid{0000-0003-0480-1991}

\author{Josiah Poon}
\affiliation{
    \institution{The University of Sydney}
    \city{Sydney}
    \state{NSW}
    \postcode{2006}
    \country{Australia}
}
\email{josiah.poon@sydney.edu.au}
\orcid{0000-0003-3371-8628}

\author{Soyeon Caren Han}
\authornote{Corresponding author.}
\affiliation{
    \institution{The University of Melbourne}
    \city{Melbourne}
    \state{VIC}
    \postcode{3052}
    \country{Australia}
}
\email{caren.han@unimelb.edu.au}
\orcid{0000-0002-1948-6819}

\renewcommand{\shortauthors}{Rina Carines Cabral, Siwen Luo, Josiah Poon, and Soyeon Caren Han}

\begin{abstract}
The significance of mental health classification is paramount in contemporary society, where digital platforms serve as crucial sources for monitoring individuals' well-being. However, existing social media mental health datasets primarily consist of text-only samples, potentially limiting the efficacy of models trained on such data.
Recognising that humans utilise cross-modal information to comprehend complex situations or issues, we present a novel approach to address the limitations of current methodologies. In this work, we introduce a \textbf{M}ultimodal and \textbf{M}ulti-Teacher Knowledge Distillation model for \textbf{M}ental \textbf{Health} Classification, leveraging insights from cross-modal human understanding.
Unlike conventional approaches that often rely on simple concatenation to integrate diverse features, our model addresses the challenge of appropriately representing inputs of varying natures (e.g., texts and sounds). To mitigate the computational complexity associated with integrating all features into a single model, we employ a multimodal and multi-teacher architecture. By distributing the learning process across multiple teachers, each specialising in a particular feature extraction aspect, we enhance the overall mental health classification performance. Through experimental validation, we demonstrate the efficacy of our model in achieving improved performance.\footnote{Code available at \url{https://github.com/adlnlp/3mhealth}}
\end{abstract}

\ccsdesc[300]{Information systems~Decision support systems}
\ccsdesc[500]{Computing methodologies~Information extraction}

\keywords{mental health classification; knowledge distillation; multimodal}



\maketitle

\section{Introduction}
\label{sec:introduction}
Mental health is a critical aspect of individual well-being, influencing both personal lives and societal structures \cite{garg2023mental}. Despite advancements in mental healthcare, not everyone with mental health concerns actively seeks professional help. The widespread use of social media platforms, such as Twitter and Reddit, has opened avenues for detecting mental health issues by analysing text-oriented posts. This shift towards online expression has prompted research into text-based mental health classification, focusing on identifying the presence and categories of mental health concerns within social media posts.
Recent studies in mental health classification from social media content have embraced diverse components, ranging from historical posts and conversation trees to social graphs and user metadata \cite{cao-2019-latent,cao-2022-building,lin-2020-sensemood,sawhney-2021-phase,sawhney-2021-towards}. However, the availability of these additional sources varies due to data privacy restrictions or user preferences, introducing challenges in research and system reproducibility. 
In light of these challenges, our research addresses the limitations of existing methodologies by focusing on the analysis of text-only social media posts, a fundamental and universally available component. While semantic pre-trained textual embedding from text-only input may capture explicit emotional words related to mental health, they may fall short in capturing less explicit emotions, limiting their robustness. For instance, some textual posts may lack explicit emotional language yet imply an unhealthy mental state.
Recognising the potential shortcomings of text-only datasets, we introduce a novel approach to mental health classification through a Multimodal and Multi-Task Knowledge Distillation Model. Inspired by human comprehension strategies that involve multimodal information integration, our model leverages insights from multimodal human understanding to enhance the efficacy of mental health risk detection.
Our approach introduces a new acoustic modality feature generated from original textual posts,
motivated by the proven effectiveness of vocal biomarkers in indicating psychological distress and other medical conditions \cite{iyer-2022-using}. This would create a new modality from text-only input for unimodal text-based mental health risk detection. Simultaneously, we also incorporate emotion-enriched features as additional information. Instead of integrating all modalities into one model, we employ a multimodal and multi-teacher architecture to address the computational complexity of integrating diverse features. This approach distributes the learning process across multiple teachers, each specialising in a particular feature extraction aspect. 
To the best of our humble knowledge, there have been no attempts to create a new modality from text-only input for the unimodal text-based mental health risk detection tasks. Additionally, we propose a new multimodality knowledge distillation model for the mental health risk detection domain.

\section{Related Works}
\subsection{Mental Health Classification}
Recent studies in mental health classification from social media content have incorporated diverse social media components. 
These components encompass various elements, including historical posts, conversation trees, social and interaction graphs, user or post metadata information, and profile pictures or posted images \cite{cao-2019-latent,cao-2022-building,lin-2020-sensemood,sawhney-2021-phase,sawhney-2021-towards}.
However, those additional sources will not always be available in the dataset due to data privacy restrictions or user preferences.  This complicates research reproducibility since each study selects features based on what social media components are available to them. 
Our research focuses on exploring mental health detection by analysing only social media textual posts, which is a compulsory component of text-based social media posts related to mental health. 
Based on the textual aspect, existing studies have worked on frequency- or score-based emotion features \cite{aragon-2023-detecting,zogan-2022-explainable}. More recent works fine-tuned contextual embeddings on emotion-based tasks to use as emotion features \cite{lara-2021-deep,sawhney-2021-phase}. These studies mainly focus on identifying and matching one type of emotion to each word or entire textual content. On the other hand, our research highlights the complexity of human emotions wherein a single word could be associated with multiple types of emotions by integrating the emotion-enriched features generated through a multi-label, corpus-based representation learning framework. 
In addition, we propose a new way to include acoustic features generated by original textual posts in this task, motivated by the effectiveness of vocal biomarkers in psychological distress and other medical condition indications \cite{iyer-2022-using}. This simultaneously processes the emotion-enriched features as additional information. To the best of our knowledge, there have been no attempts to create a new modality from the text-only input to achieve unimodal text-based mental health risk detection tasks.

\subsection{Multi-teacher Knowledge Distillation}
Other multimodal social media-based mental health detection studies mainly integrate modalities through concatenation of features \cite{han2020victr,bucur-2023-its} or a joint encoder \cite{anshul-2024-multimodal} which doesn't represent the modalities properly. To integrate the multimodal knowledge efficiently, we design our model using knowledge distillation to compress a complex and large multimodal integration model into a smaller and simpler one while still retaining the accuracy and performance of the resultant model.
Knowledge distillation \cite{hinton2015distilling} involves transferring knowledge from a teacher model to a student model, commonly applied to compress large models by mapping intermediate layer outputs \cite{chen2021cross,jiao2020tinybert} or minimising KL-divergence in class distribution \cite{mirzadeh2020improved}. Traditionally, knowledge distillation is used within the same modality. However, recent approaches extend it to different modalities \cite{yang2021cross,ni2022cross,li2020towards}. Some studies explore collaborative learning with multiple teachers for improved compression 
such as \cite{wu2021one} and \cite{pham2023collaborative} in language and vision models. \cite{tan2018multilingual} focuses on multilingual language translation, and \cite{vongkulbhisal2019unifying} applies multi-teacher knowledge distillation to unify classifiers trained on distinct data sources. 
Inspired by this, we propose a new Multimodal Multi-Teacher Knowledge Distillation framework for mental health risk detection.

\begin{figure}[t]
    \centering
     \includegraphics[scale=0.20]{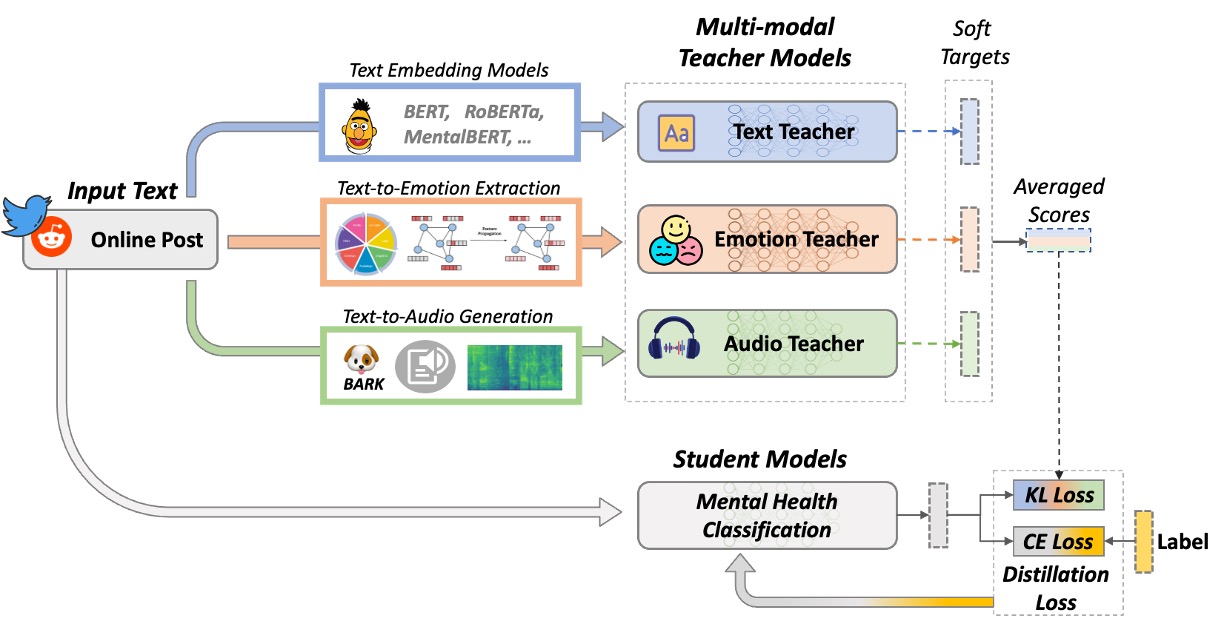}
     \caption{Architecture of 3M-Health: Multimodal Multi-teacher Knowledge Distillation for Mental Health Detection.}
     \label{fig:OverallArchitecture}
    \hfill
\end{figure}

\section{3M-Health}
In this section, we introduce our \textbf{M}ultimodal \textbf{M}ulti-teacher knowledge distillation model for \textbf{M}ental \textbf{Health} detection, \textbf{3M-Health}. Figure \ref{fig:OverallArchitecture} illustrates the overall architecture.
This model consists of three distinct teacher models, each focusing on different modalities to independently learn diverse aspects of features crucial for interpreting mental health-related posts. The acquired features from multimodal teachers serve as a valuable source of knowledge for instructing the student model by utilising the average output distribution of the teacher models as soft targets. 
We introduce three essential multimodal teacher models for mental health risk detection, including 1) a text-based teacher for understanding semantic aspects from input texts, 2) an emotion-based teacher for interpreting emotion aspects from input texts, and 3) an audio-based teacher for discerning emotions conveyed through audio sounds.

\subsection{Multimodal Multi-Teacher Construction}
This section articulates each teacher model's objective and construction process. Teacher fine-tuning can be found in Section \ref{fine-tuning}.

\subsubsection{Text-based Teacher}
The text-based teacher aims to teach contextual semantic comprehension of mental health-related textual posts. We leverage pre-trained large language models (PLMs) since contextualised embeddings from PLMs represent different meanings based on the context (e.g. \textit{blue} means \textit{a kind of the colour}, but \textit{gloomy} in other emotional contexts).  More specifically, some words may have opposite meanings in the medical domain (e.g., \textit{positive} usually means something good but often refers to the presence of a specific condition, which is typically not a desirable outcome). Inspired by this, we explore several general (BERT \cite{devlin-2018-bert}, RoBERTa \cite{liu-2019-roberta}) and medical/mental-health specific PLMs (MentalBERT \cite{ji-2021-mentalbert}, and ClinicalBERT \cite{wang-2023-optimized}) in the experiment.

\subsubsection{Emotion-based Teacher}
\label{sub:emoteacher}
The emotion-based teacher aims to teach emotional aspects from the input text of mental health-related posts. It is initialised with representations derived from a graph-based model \cite{cabral-2024-mm}, which produces emotion-enriched word representations by thoroughly incorporating global and local relationships among posts and all the words within those posts. To do so, we first obtain a multi-label emotion class indicating \textit{anger}, \textit{disgust}, \textit{fear}, \textit{sadness}, \textit{surprise}, \textit{negative}, and \textit{other}\footnote{Positive sentiment/emotions are grouped into \textit{other} to focus on the different negative emotion on mental health-related text.} for each post using the SenticNet7 lexicon \cite{cambria-2022-senticnet}\footnote{\url{https://sentic.net/downloads/}}, mapping identified words to their corresponding emotion types. This emotion lexicon consists of terms $K = \{k_1, ..., k_q\}$ associated with one or more emotion types from $EM = \{em_1, ..., em_r\}$. For each word $W = \{w_1, ..., w_p\}$ in a post, we assign $EM_{k_j}$ to $w_i$ whenever $w_i = k_j$ in $K$ in this document. Consequently, each post is associated with a multi-label class $EM_d = \{EM_{w_1} \cup EM_{w_2} \cup ... EM_{w_p}\}$.
Subsequently, we construct a graph $G=(V,E,A)$ representing all posts and their word tokens, where $V$ is the set of all post nodes and token nodes tokenised through wordpiece tokenisation with emoticon preservation\footnote{To further preserve and integrate emotions in the posts, emoticons and emojis are added to the tokeniser vocabulary.}. Here, $E$ encompasses token-token edges $E_{w_i, w_j}$, token-post edges $E_{w_i, d_j}$, and post-post edges $E_{d_i, d_j}$, while $A$ specifies weights between related nodes\cite{han-2022-understanding}. Post node and token node representations are initialised with the [CLS] embedding and the minimum of contextualised token embeddings from pre-trained BERT word embeddings, respectively. Edge values are determined by Pointwise Mutual Information (PMI) for $E_{w_i, w_j}$, Term Frequency-Inverse Document Frequency (TF-IDF) for $E_{w_i, d_j}$, and Jaccard similarity for $E_{d_i, d_j}$. Utilising these initialised representations and edge values, a two-layer Graph Convolutional Neural Network \cite{kipf-2016-semi} (GCN) is trained with ReLu for the multi-label emotion classification task based on the SenticNet7. The updated second-layer hidden states are extracted and used as initial weights for fine-tuning BERT on the same multi-label emotion classification task to comprehend the associated emotions in the posts further. The updated word embeddings are extracted as the multi-emotion contextual embeddings.

\subsubsection{Audio-based Teacher}
Existing social media mental health datasets primarily consist of text-only samples, so the two teachers mentioned before are purely based on the text. To address the need for a more comprehensive understanding of complex mental health and emotional contexts, we propose the integration of multimodality information. According to research \cite{collins2023conveying}, individuals can more accurately interpret the emotions of others through listening rather than observing facial expressions/body language or reading written text. Drawing inspiration from this insight, we introduce an audio-based teacher to enhance knowledge distillation, enabling the interpretation of emotions in mental health posts through sound.
To achieve this, we first employ Bark, a pre-trained text-to-audio model, to generate corresponding audio for each post as Bark can capture emotional sounds detected from the text (e.g. \textit{[laughs]}, \textit{[gasps]} and \textit{``...”} for hesitations)\footnote{A theoretical and practical comparison of text-to-audio generation APIs and a list of Bark's sound cues in Section \ref{sec:audioapi}.}. Note that Bark can generate only 13 seconds of audio. Hence, we tokenise each post at the sentence level, generate audio for each sentence, and then aggregate these audio segments into a complete audio representation for the entire post. Particularly long sentences or texts that do not have punctuation are further split into a maximum of 45 tokens.

\subsection{Multimodal Multi-Teacher Fine-tuning}
\label{fine-tuning}
Researchers \cite{jiao2020tinybert,wu2021one} emphasise the significance of distilling knowledge from the hidden states of a teacher model for effective student instruction. In this section, we describe the independent fine-tuning process of each teacher model.

We fine-tuned pre-trained language models 
for the mental health classification task with labels $C = \{c_1,c_2, . . . ,c_{|C |} \}$ to build the text-based teachers. This process enables the teacher to learn the nuances of mental health-related contexts within each dataset.
For the emotion-based teacher, following the generation of emotion-rich representations for each post and its words, these embeddings serve as inputs for fine-tuning a Multi-Layer Perceptron (MLP) for mental health classification, operating over the labels $C$.
We employ the Audio Spectrogram Transformer \cite{gong-ast-2021} (AST) for the audio-based teacher to classify each generated audio into mental health risk classes. AST is a transformer-based model that takes a sequence of audio spectrogram patches as inputs. An audio waveform is first converted into a 128x100t spectrogram based on a sequence of 128-dimensional log Mel filterbank features computed with a 25ms Hamming window every 10ms. Such a spectrogram is then split into a sequence of N 16x16 patches of images with an overlap of 6 in both time and frequency dimensions. A special token [CLS] is added to the beginning of the sequence of spectrogram patches. After passing through transformer encoder layers, the [CLS] embedding is fed into a linear layer with sigmoid activation to classify mental health risk class labels $C$.

Every teacher model is individually constructed and fine-tuned to facilitate optimal learning. We are aware of the concerns raised by some researchers \cite{wu2021one} highlighting the potential inconsistency in the feature space when different teachers are separately pre-trained with distinct settings and then fine-tuned independently. Based on our initial testing, co-finetuning multimodal teachers yields little improvement; in fact, it tends to result in lower performance. We speculate that integrating multimodal information may not perform optimally during co-finetuning.

\begin{table}[t]
\footnotesize
\centering
\caption{Data statistics. Durations are in a minute:second (mm:ss) format. $^\pm$SDCNL categorises suicide and depression-related posts.}
\begin{tabularx}{\columnwidth}{
    X |
    >{\centering\arraybackslash}b{0.14\columnwidth} 
    >{\centering\arraybackslash}b{0.14\columnwidth} 
    >{\centering\arraybackslash}b{0.14\columnwidth} 
    >{\centering\arraybackslash}b{0.14\columnwidth} 
    >{\centering\arraybackslash}b{0.14\columnwidth}}
    \toprule
        & \textbf{TwitSuicide} 
        & \textbf{DEPTWEET}
        & \textbf{IdenDep} & \textbf{SDCNL} \\
    \midrule
    Task & Suicide 
    & Depression
    & Depression & Suicide$^\pm$ \\
    Platform & Twitter 
        & Twitter
        & Reddit & Reddit \\
    \hline
    Num. Classes & 3 
        & 4
        & 2 & 2 \\
    Total Samples & 660 
        & 5128
        & 1841 & 1895 \\
    \hline
    Evaluation & 10-fold 
        & 60/20/20
        & 10-fold & 80/20 \\
    Train/Val & -
        & 4,102 
        & - & 1,516 \\
    Test & - 
        & 1,026
        & - & 379 \\
    \hline
    Length & 13-147 
        & 1-926
        & 11-17,641 & 13-24,590 \\
    Avg. Length & 90.32
        & 163.28 
        & 1,127.57 & 936.76 \\
    Words & 3-31 
        & 101
        & 3,477 & 4,411 \\
    Avg. Words & 16.85 
        & 28.15
        & 215.1 & 178.53 \\
    \hline
    
    Min Duration
        & 00:01.903
        & 00:01.250
        & 00:02.900
        & 00:02.463
        \\
    Max Duration
        & 00:32.853
        & 00:56.596
        & 22:07.740
        & 28:09.546
        \\
    Avg. Duration
        & 00:11.545
        & 00:17.860
        & 01:45.193
        & 01.27.568
        \\
    \bottomrule
\end{tabularx}
\label{tbl:datastatistics}
\end{table}

\begin{figure}[t]
\captionsetup[sub]{justification=centering}
    \centering

    \begin{subfigure}[t]{0.48\columnwidth}
         \centering
         \includegraphics[scale=0.40]{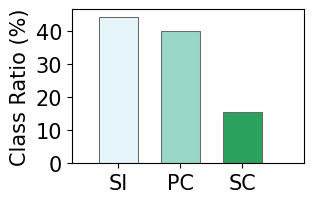}
         \caption{\textbf{TwitSuicide}}
         \label{fig:ClassDistTwitter}
    \end{subfigure}
    \begin{subfigure}[t]{0.48\columnwidth}
         \centering
         \includegraphics[scale=0.40]{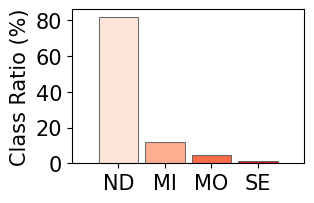}
         \caption{\textbf{DEPTWEET}}
         \label{fig:ClassDistDEPTWEET}
    \end{subfigure}
    \begin{subfigure}[t]{0.48\columnwidth}
         \centering
         \includegraphics[scale=0.40]{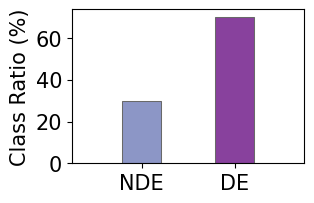}
         \caption{\textbf{IdenDep}}
         \label{fig:ClassDistIdenDep}
    \end{subfigure}
    \begin{subfigure}[t]{0.48\columnwidth}
         \centering
         \includegraphics[scale=0.40]{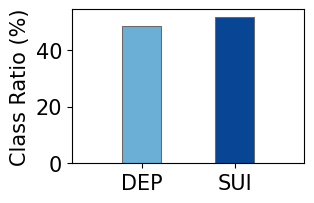}
         \caption{\textbf{SDCNL}}
         \label{fig:ClassDistSDCNL}
    \end{subfigure}
    \caption{Class distribution. For (a) TwitSuicide, SI: Safe to Ignore; PC: Possibly Concerning; SC: Strongly Concerning. For (b) DEPTWEET, ND: Non-depression; MI: Mild; MO: Moderate; SE: Severe. For (c) IdenDep, NDE: Non-depression; DE: Depression. For (d) SDCNL, DEP: Depression; SUI: Suicide.}
    \label{fig:ClassDist}
\end{figure}

\begin{figure}[t]
    \begin{subfigure}[t]{0.48\columnwidth}
         \centering
         \includegraphics[scale=0.38]{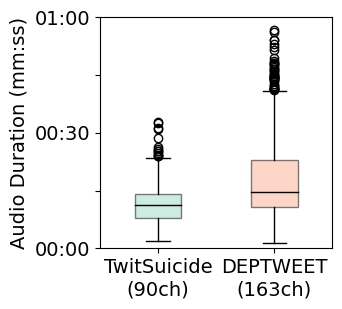}
         \caption{Twitter-based}
         \label{fig:AudioBoxTwitter}
    \end{subfigure}  
    \begin{subfigure}[t]{0.48\columnwidth}
         \centering
         \includegraphics[scale=0.38]{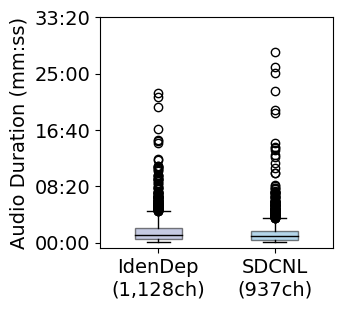}
         \caption{Reddit-based}
         \label{fig:AudioBoxReddit}
    \end{subfigure} 
    \caption{Audio length comparison. ch: character average}
     \label{fig:AudioBoxOutlier}
\end{figure}

\subsection{Multi-Teacher Knowledge Distillation}
For the student model, we use a single modality involving textual posts as input for a pre-trained BERT, which processes the sequence of tokenised words. The student model performs the mental health risk classification task over the same class labels $C$, with knowledge distilled from the text-based, emotion-based, and audio-based teacher models. To incorporate the acquired knowledge from these various multimodal sources, the student model is trained to minimise the distillation loss given by $L = L_{task} + L_{kd}$. Here, $L_{task}$ represents the cross-entropy loss between the student model's predictions and the ground truth of mental health risk categories, while $L_{kd}$ stands for the Kullback-Leibler (KL) divergence between the student model and the teacher models' predictions. Given the presence of multiple teacher models, we calculate $L_{kd}$ by averaging the predicted probability distributions from all three teacher models.

\section{Experimental Setup}
\subsection{Datasets}


\begin{table}[t]
\footnotesize
\centering
\caption{Text statistics for each class per dataset.}
\begin{tabularx}{\columnwidth}{
    X |
    >{\centering\arraybackslash}p{0.04\columnwidth} 
    >{\centering\arraybackslash}p{0.04\columnwidth} 
    >{\centering\arraybackslash}p{0.23\columnwidth} 
    >{\centering\arraybackslash}p{0.19\columnwidth} 
    }
    \toprule  
    \textbf{Class}
        & \textbf{Total}
        & \textbf{\%}
        & \textbf{Length (ave.)}
        & \textbf{Words (ave.)}
        \\
    \midrule
    \multicolumn{5}{l}{\textbf{TwitSuicide}} \\
    \midrule
    Safe to Ignore
        & 103 
        & 15.61
        & 13-139 (77.89)
        & 4-31 (15.25)
        \\
    Possibly Concerning
        & 264 
        & 40.00
        & 24-147 (88.16)
        & 4-31 (16.35)
        \\
    Strongly Concerning
        & 293 
        & 44.39
        & 13-147 (96.65)
        & 3-30 (17.86)
        \\
    \midrule
    \multicolumn{5}{l}{\textbf{DEPTWEET}} \\
    \midrule
    Non-Depressed
        & 4213 
        & 82.16
        & 1-816 (164.47)
        & 1-101 (28.08)
        \\
    Mild
        & 606 
        & 11.82
        & 4-885 (144.74)
        & 1-87 (26.38)
        \\
    Moderate
        & 232 
        & 4.52
        & 32-926 (184.95)
        & 5-99 (33.25)
        \\
    Severe
        & 77 
        & 1.50
        & 23-398 (178.81)
        & 1-62 (30.57)
        \\

    \midrule
    \multicolumn{5}{l}{\textbf{IdenDep}} \\
    \midrule
    Non-Depression
        & 548 
        & 29.77
        & 11-17641 (1546.34)
        & 1-3477 (295.75)
        \\
    Depression
        & 1293 
        & 70.23
        & 11-13803 (950.09)
        & 2-2487 (180.92)
        \\
    
    \midrule
    \multicolumn{5}{l}{\textbf{SDCNL}}\\
    \midrule
    Depression
        & 915 
        & 48.28
        & 43-16015 (1000.68)
        & 8-3200 (192.84)
        \\
    Suicide
        & 980 
        & 51.72
        & 13-24590 (977.07)
        & 2-4411 (165.16)
    \\
        
    \bottomrule 
\end{tabularx}
\label{tbl:classtextstatistics}
\end{table}

\begin{table}[t]
\footnotesize
\centering
\caption{Audio statistics for each class per dataset in a minute:second (mm:ss) format.}
\begin{tabularx}{\columnwidth}{
    >{\RaggedRight\arraybackslash}b{0.15\columnwidth} 
    X |
    >{\centering\arraybackslash}b{0.12\columnwidth}
    >{\centering\arraybackslash}b{0.12\columnwidth}
    >{\centering\arraybackslash}b{0.12\columnwidth}
    }
    \toprule  
    \textbf{Dataset}
        & \textbf{Class}
        & \textbf{Min Duration}
        & \textbf{Max Duration}
        & \textbf{Ave. Duration}
        \\
    \midrule
    \multirow{3}{*}{\textbf{TwitSuicide}}
    & Safe to Ignore
        & 00:01.903
        & 00:31.000
        & 00:12.215
        \\
    & Possibly Concerning
        & 00:02.143
        & 00:32.853
        & 00:11.393
        \\
    & Strongly Concerning
        & 00:01.943
        & 00:24.760
        & 00:10.280
        \\
    \midrule
    \multirow{4}{*}{\textbf{DEPTWEEET}}
    & Non-Depressed
        & 00:01.250
        & 00:56.200
        & 00:17.210
        \\
    & Mild
        & 00:02.230
        & 00:56.596
        & 00:15.413
        \\
    &Moderate
        & 00:04.460
        & 00:47.770
        & 00:18.958
        \\
    & Severe
        & 00:02.250
        & 00:40.160
        & 00:17.865
        \\

    \midrule
    \multirow{2}{*}{\textbf{IdenDep}}
    & Non-Depression
        & 00:02.900
        & 22:07.740
        & 02:20.941
        \\
    & Depression
        & 00:03.100
        & 21:28.643
        & 01:30.420
        \\
    
    \midrule
    \multirow{2}{*}{\textbf{SDCNL}}
    & Depression
        & 00:05.756
        & 25:58.966
        & 01:33.802
        \\
    & Suicide
        & 00:02.463
        & 28.09.546
        & 01:21.747
    \\
        
    \bottomrule 
\end{tabularx}
\label{tbl:classaudiostatistics}
\end{table}

We evaluate our proposed model using four publicly available datasets related to mental health on social media. Table \ref{tbl:datastatistics} and Figure \ref{fig:ClassDist} provide a summary of statistics and class distribution.

The \textbf{TwitSuicide Dataset}\footnote{Data available upon request.} \cite{long-2022-aqualitative} replicates the data collection, processing, and annotation methods of \cite{odea-2015-detecting}. A sample of 660 tweets is annotated into three risk levels.
The \textit{Strongly Concerning} (SC) class is assigned to posts with a convincing display of severe suicidal ideation, while \textit{Safe to Ignore} (SI) shows no evidence of suicide risk. If it doesn't fall into other categories, a post remains in the \textit{Possibly Concerning} (PC) class.
\textbf{DEPTWEET}\footnote{\url{https://github.com/mohsinulkabir14/DEPTWEET}} \cite{kabir-2023-deptweet} is collected from Twitter using seed terms based on the Patient Health Questionnaire (PHQ-9). 
The dataset comprises 40,191 tweets; however, only 5,128 tweets were retrieved during this study. The labels include \textit{Non-Depressed} (ND), \textit{Mild} (MI), \textit{Moderate} (MO), and \textit{Severe} (SE), maintaining an imbalanced class distribution, with around 80\% labelled as ND and less than 2\% SE.
The \textbf{Identifying Depression Dataset}\footnote{\url{https://github.com/Inusette/Identifying-depression}} (IdenDep) \cite{pirina-2018-identifying} consists of 1,841 Reddit posts, with ``depression indicative” (DE) posts sourced from the Depression subreddit and non-depressive (NDE) posts from the ``family” and ``friendship advice” subreddits. No further manual check was done on the samples, increasing the probability of false negatives.
The \textbf{SDCNL Dataset}\footnote{\url{https://github.com/ayaanzhaque/SDCNL}} \cite{haque-2021-deep} involves distinguishing between Reddit suicide-related and depression-related posts. The dataset contains 1,895 nearly balanced posts labelled as \textit{Suicide} (SUI) or \textit{Depression/Not Suicide} (DEP) based on their subreddit.
In accordance with \cite{benton-2017-ethical}, all posts are de-identified before any analysis, audio generation, and model training.



We provide a detailed breakdown of text and audio statistics in Tables \ref{tbl:classtextstatistics} and \ref{tbl:classaudiostatistics} to provide more information regarding the nature of each class, which may influence model learning and performance. Notably, DEPTWEET and IdenDep datasets have highly skewed data, with 82.16\% and 70.23\% on a single class, respectively. Figure 
\ref{fig:AudioBoxOutlier} illustrates the differences in the generated audio in terms of duration. The Reddit-based datasets, IdenDep and SDCNL, are significantly longer than the Twitter-based datasets, possibly providing more auditory information inferred from the textual posts.

\subsection{Text-to-Audio Generators}
\label{sec:audioapi}

In order to generate the best possible audio to represent each textual post in our benchmark datasets, we performed a theoretical and practical comparison between five publicly accessible text-to-speech and text-to-audio generative APIs.

    \textbf{Tacotron2}\footnote{\url{https://github.com/NVIDIA/tacotron2}} \cite{shen-2018-natural} uses a recurrent neural network architecture to predict mel spectrogram sequences from text followed by a modified WaveNet vocoder.     
    \textbf{SpeechT5}\footnote{\url{https://github.com/microsoft/SpeechT5}}     \cite{ao-2022-speecht5} unifies modalities with a shared encoder-decoder architecture that uses cross-modal vector quantisation for speech and text alignment.     
    \textbf{SpeechBrain}\footnote{\url{https://github.com/speechbrain/speechbrain/}} \cite{ravanelli-2021-speechbrain} is a speech toolkit offering various speech related tasks. Their text-to-speech model is based on Tacotron2 but is trained further on the LJSpeech \cite{ito-2017-ljspeech} and LibriTTS \cite{zen-2019-libritts} datasets. 
    \textbf{Balacoon}\footnote{\url{https://huggingface.co/balacoon/tts}} packages offer lightweight and fast text analysis and speech generation going against larger but slower TTS models. It sacrifices multi-speaker and multi-lingual features for lightning-fast speed on the CPU. The detailed model architecture was not publicly available at the time of this paper's writing.     
    \textbf{Bark}\footnote{\url{https://github.com/suno-ai/bark}} has a GPT-based architecture using a quantised audio representation that does not require the use of phonemes allowing it to generalise beyond speech, thus making it a text-to-audio model. 

Upon comparison of the five generators, we use Bark due to the expressiveness of the audio generated by the model. While the other models suffer from a robotic delivery of the generated speech, not verbalising numerical figures, and reading of emoticons as individual punctuations (e.g. ``\textgreater:\textbar” as \textit{greater than, semicolon, pipeline}), Bark produces the most naturally sounding audio recognising textual markers like ``,” for pauses, ``--” and ``...” for hesitations, capitalisation for emphasis (e.g. \textit{goodbye} vs. \textit{GOODBYE}), and sentence punctuations for produce tonal shifts (e.g. \textit{huh?} vs \textit{huh!}). Bark also verbalises non-speech sounds such as [laughter], [laughs], [sighs], [music], [gasps], [clears throat], \textit{haha}, \textit{uhm}, \textit{waaah}, and \textit{ooh}. Bark's ability to infer and convey emotions from only an input text would be valuable to our mental health risk detection model since it can provide additional emotional cues from the generated sound.

\subsection{Baselines and Metrics}
\label{sec:baselines}
We assess the performance of our model by comparing it to previously published results using post-only\footnote{In contrast to studies incorporating other components such as posted images or user network and activity.} and post-level classification on the same datasets, employing identical class labels and similar evaluation setups.  We use results reported in the following studies as our baselines: \textbf{Bi-LSTM Char\texttt{+}Word} \cite{long-2022-aqualitative} for TwitSuicide; 
\textbf{MLP} \cite{tadesse-2019-detection}, and \textbf{EAN} \cite{ren-2021-depression} for IdenDep; and \textbf{GUSE-DENSE} \cite{haque-2021-deep}, and  \textbf{AugBERT+LR} \cite{ansari-2021-data} for SDCNL. For the DEPTWEET dataset, we use the published \textbf{DistilBERT} code from \cite{kabir-2023-deptweet} to replicate baseline results for the retrieved dataset. 
In addition, we provide strong baselines from fine-tuning state-of-the-art PLMs: \textbf{BERT} \cite{devlin-2018-bert}, \textbf{RoBERTa} \cite{liu-2019-roberta}, \textbf{MentalBERT}, and \textbf{MentalRoBERTa} \cite{ji-2021-mentalbert}. 
All PLM baselines follow the training setup used by \cite{long-2022-aqualitative} with a batch size of 8 and a learning rate of 1e-04 trained for three epochs. 
Given the class imbalance, we evaluate our system based on macro F1 (F1m) and weighted F1 (F1w) scores, followed by accuracy and class F1 scores.

\subsection{Implementation Details}

We evaluate our model following established evaluation setups from previous literature using the same datasets on the same classification task setup for fair benchmark comparisons. We use 10-fold cross-validation for TwitSuicide and IdenDep, while a train/test split is used for SDCNL and DEPTWEET. Original data splits are retained when provided; otherwise, the data is randomly split (Table \ref{tbl:datastatistics}). When none is given, 10\% of the training set is used for validation.

Hyperparameter tuning is done per dataset and model setup using Optuna\footnote{\url{https://optuna.org/}}
optimising weighted F1 scores. 
Detailed search space and best-found hyperparameters may be found in the Supplementary Material.
Text-based teachers are trained using ReLu and a max length of 256.
Audio-based teachers are trained for 25 epochs with a 5-epoch early stop, a 512 max length, and using the ReduceLRonPlateau scheduler.
All inputs for the AST model are normalised to zero mean and 0.5 standard deviation.
Emotion-based teachers are trained for 100 epochs with a 10-epoch early stop and a 256 max length.
The student models are tuned and trained with distilled knowledge from the fine-tuned teachers using a max length of 256.
All tuning was done using a 90:10 split and was conducted separately from the final model construction.
All models are trained using an Adam optimiser on an NVIDIA TITAN RTX machine.

\section{Results}
\subsection{Overall Performance}
\label{sec:OverallResults}

\begin{table}[t]
\footnotesize
\centering
\caption{Overall results using all three teacher modalities (\textbf{Ours (All)}) and the best partial teacher combination (\textbf{Ours (Best Partial Combination)}) against baselines. Class abbreviation definitions may be found in the Figure \ref{fig:ClassDist} caption. We present a full teacher combination ablation study in Table \ref{tbl:ablationstudyteachers}. $^\dagger$ indicates replicated results. \textbf{Bold} face indicates best score while second best are \underline{underlined}.}
\begin{tabularx}{\columnwidth}{
    X |
    >{\centering\arraybackslash}b{0.05\columnwidth}
    >{\centering\arraybackslash}b{0.05\columnwidth}
    >{\centering\arraybackslash}b{0.05\columnwidth} |
    >{\centering\arraybackslash}b{0.05\columnwidth} 
    >{\centering\arraybackslash}b{0.05\columnwidth} 
    >{\centering\arraybackslash}b{0.05\columnwidth} 
    >{\centering\arraybackslash}b{0.05\columnwidth} 
    }
    \toprule
    &\multicolumn{3}{c|}{\textbf{Overall Performance}} 
    &\multicolumn{4}{c}{\textbf{Breakdown F1 Scores}}\\
    \toprule  
    \textbf{TwitSuicide} 
        & 
            \textbf{Acc} 
        & 
            \textbf{F1m}
        & 
            \textbf{F1w}
        & 
            \textbf{(SC)}
        & 
            \textbf{(PC)}
        & 
            \textbf{(SI)}
        \\
    \hline
    \citet{long-2022-aqualitative} 
        & 56.67
        & -
        & -
        & 40.00
        & 50.00
        & \underline{66.00}\\
    \hline
    BERT 
        & 57.58
        & 53.60
        & 57.25
        & 40.00
        & 57.00
        & 64.00
        \\
    RoBERTa
        & 55.45
        & 50.61
        & 54.43
        & 37.18
        & 51.63
        & 63.03
        \\
    MentalBERT 
        & 57.73
        & 52.57
        & 57.39
        & 35.23
        & 56.65
        & 65.84
        \\    
    MentalRoBERTa
        & 55.91
        & 51.49
        & 55.60
        & 41.62
        & 53.05
        & 62.81
        \\
    \hline
    \textbf{Ours (Text\&Emo)} 
        & \textbf{65.76}        
        & \textbf{61.96}
        & \textbf{65.46}
        & \underline{49.72}
        & \textbf{62.34}
        & \textbf{73.81}
        \\
    \textbf{Ours (All)} 
        & \underline{61.21}
        & \underline{59.64}
        & \underline{61.23}
        & \textbf{54.17}
        & \underline{59.50}
        & 65.26
        \\
        
    \toprule
    \textbf{DEPTWEET} & \textbf{Acc} & \textbf{F1m}  & \textbf{F1w} 
    & \textbf{(SE)} & \textbf{(MO)} & \textbf{(MI)} & \textbf{(ND)}  \\
    \hline   
    \citet{kabir-2023-deptweet}$^\dagger$
        & 79.75 
        & 38.59
        & 78.89 
        & 17.65 
        & \underline{21.98} 
        & 25.43 
        & 89.29 
        \\
    \hline
    BERT 
        & 81.89
        & 40.40
        & 80.21
        & \underline{36.36} 
        & 00.00 
        & 34.34
        & 90.90 
        \\    
    RoBERTa
        & 82.18
        & 32.04
        & 80.14
        & 00.00
        & 00.00
        & \underline{36.77}
        & 91.41
        \\
    MentalBERT
        & \underline{83.54}
        & 36.04
        & 79.72
        & 26.09
        & 00.00
        & 26.67
        & 91.41
        \\
    MentalRoBERTa
        & 78.48
        & 36.60
        & 78.62
        & 22.22
        & 00.00
        & 34.91
        & 89.28
        \\
    \hline
    \textbf{Ours (Text\&Emo)}
        & \textbf{84.03} 
        & \textbf{46.43}
        & \textbf{83.09} 
        & \textbf{41.38} 
        & 12.31
        & \textbf{39.07} 
        & \textbf{92.95} 
        \\
    \textbf{Ours (All)}
        & 82.77 
        & \underline{46.20}
        & \underline{82.61}
        & 26.09 
        & \textbf{34.34} 
        & 32.32 
        & \underline{92.04}
        \\
        
    \toprule    
    \textbf{IdenDep} 
        & \textbf{Acc}         
        & \textbf{F1m} 
        & \textbf{F1w} 
        & \textbf{(DE)}
        & \textbf{(NDE)}
        \\
    \hline   
    \citet{tadesse-2019-detection} 
        & 91.00 
        & - 
        & -
        & 
        93.00
        & 
        -
    \\
    \citet{ren-2021-depression} 
        & 91.30
        & - 
        & - 
        & 93.98
        & -
        \\
    \hline
    BERT
        & 88.65
        & 85.23
        & 88.10
        & 92.34
        & 78.12
        \\
    RoBERTa
        & 87.18
        & 82.85
        & 86.34
        & 91.47
        & 74.24
        \\
    MentalBERT
        & 89.63
        & 86.71
        & 89.23
        & 92.93
        & 80.49
        \\
    MentalRoBERTa 
        & 90.11 
        & 87.70
        & 89.91
        & 93.15
        & 82.26
        \\
    \hline
    \textbf{Ours (Text\&Audio)} 
        & \textbf{94.30} 
        & \textbf{93.10}
        & \textbf{94.26} 
        & \textbf{95.97}
        & \textbf{90.23}
        \\
    \textbf{Ours (All)}
        & \underline{93.92}
        & \underline{92.58}
        & \underline{93.85}
        & \underline{95.73}
        & \underline{89.43}
        \\
        
    \toprule    
    \textbf{SDCNL} 
        & \textbf{Acc} 
        & \textbf{F1m} 
        & \textbf{F1w}
        & 
        \textbf{(SUI)}
        & 
        \textbf{(DEP)}
        \\
    \hline   
    \citet{haque-2021-deep} 
        & 72.24 
        & - 
        & - 
        & 
        73.61
        & 
        -
        \\
    \citet{ansari-2021-data} 
        & - 
        & - 
        & - 
        & 76.00
        & -
        \\
    \hline
    BERT
        & 67.02
        & 66.57
        & 66.64
        & 70.45
        & 62.69
        \\
    RoBERTa
        & 70.97
        & 70.63
        & 70.69
        & 73.81
        & 67.46
        \\
    MentalBERT
        & 69.39
        & 69.21
        & 69.26
        & 71.57
        & 66.86
        \\
    MentalRoBERTa  
        & 72.30
        & 72.10
        & 72.14
        & 74.45
        & 69.74
        \\
    \hline
    \textbf{Ours (Text\&Audio)} 
        & \textbf{76.52} 
        & \textbf{76.50}
        & \textbf{76.51} 
        & \underline{77.12}
        & \textbf{75.88}
        \\
    \textbf{Ours (All)}
        & \underline{75.20}
        & \underline{74.84}
        & \underline{74.90}
        & \textbf{77.83}
        & \underline{71.86}
        \\
        
    \bottomrule
\end{tabularx}
\label{tbl:overallresults}
\end{table}

\begin{figure}[t]
    \centering
    \begin{subfigure}[t]{\columnwidth}
         \centering
         \includegraphics[scale=0.37]{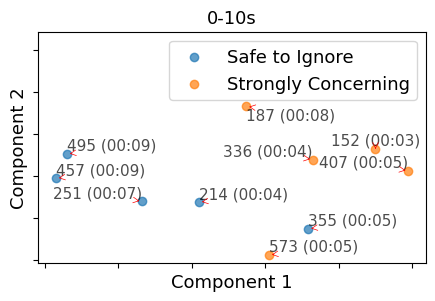}
         \includegraphics[scale=0.37]{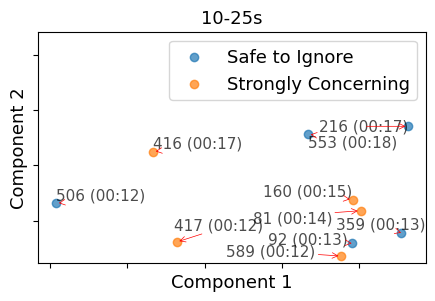}
         \caption{TwitSuicide}
         \label{fig:AudioScatterTwitter}
    \end{subfigure}    
    \begin{subfigure}[t]{\columnwidth}
         \centering
         \includegraphics[scale=0.37]{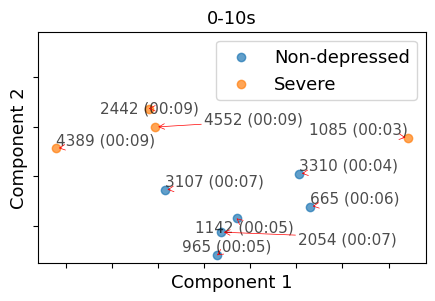}
         \includegraphics[scale=0.37]{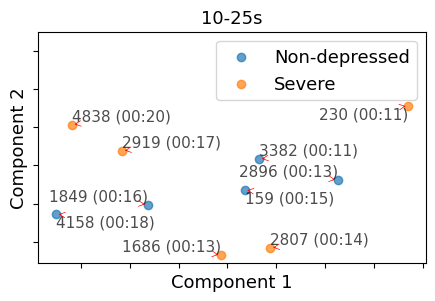}
         \caption{DEPTWEET}
         \label{fig:AudioScatterDEPTWEET}
    \end{subfigure}  
    \begin{subfigure}[t]{\columnwidth}
         \centering
         \includegraphics[scale=0.37]{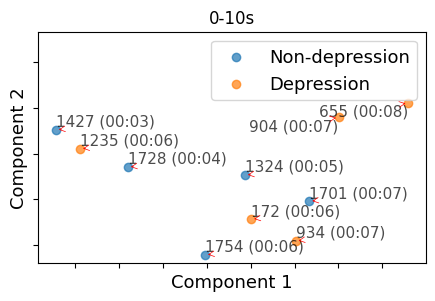}
         \includegraphics[scale=0.37]{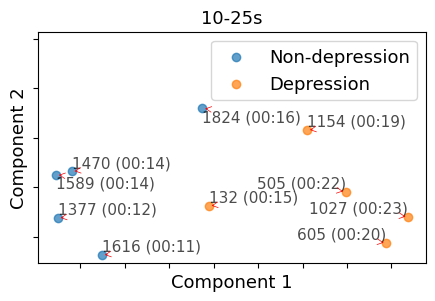}
         \caption{IdenDep}
         \label{fig:AudioScatterIdenDep}
    \end{subfigure}  
    \begin{subfigure}[t]{\columnwidth}
         \centering
         \includegraphics[scale=0.37]{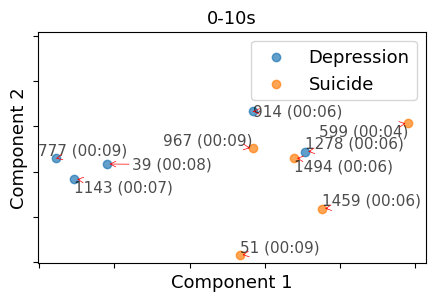}
         \includegraphics[scale=0.37]{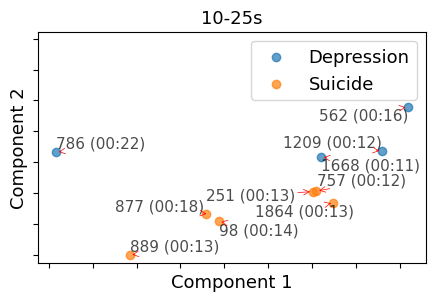}
         \caption{SDCNL}
         \label{fig:AudioScatterSDCNL}
    \end{subfigure}  
    \caption{Audio analysis using PCA on spectrogram images of audio samples grouped by a maximum of 10s (left) and 10-25s (right). Each sample is labelled with an ID for reference to corresponding texts provided in the Supplementary Material.}
    \label{fig:AudioAnalysis}
\end{figure}

We compare our model with fine-tuned PLM baselines and several published baselines that use the same mental health detection datasets and evaluation setup. Note that we select post-only mental health detection models as mentioned in Section \ref{sec:introduction} and \ref{sec:baselines}. We comprehensively evaluate the overall performance and class performance in Table \ref{tbl:overallresults}.

Overall, our model outperforms all baselines on all four benchmark datasets. What should be noted is that our model does not have to be trained with all three different teachers to achieve the best results. 
As illustrated in Table \ref{tbl:overallresults}, we have four datasets, the initial two originating from Twitter and the latter from Reddit. Our model demonstrates superior performance, even with partial teacher combinations. The datasets from Twitter produce the best results with the combination of Text and Emotion, whereas the Reddit-based datasets perform the best with Text and Audio Knowledge Distillation. Note that we detail the efficacy of each modality teacher combination on different datasets in Section \ref{sec:AblationStudy}. 
In addition, our model trained with all three modalities still outperforms the other baseline models in most cases and shows greater performance in identifying certain classes. Especially for the \textit{Moderate} (MO) class of the DEPTWEET dataset, our model trained with all three modalities achieves a 34.34 F1 score, while our model trained with partial modalities only achieves a 12.31 F1 score. All the other pre-trained baseline models fail to recognise the MO class. Hence, we can conclude that learning from different modality teachers helps our model achieve much better performances than the baseline models that learned from only textual inputs. Such improvement is more noticeable on datasets with shorter texts. Specifically, our model’s best performance is 8.36\% and 8.07\% higher than the best-performing baseline model on the macro F1 and weighted F1, respectively, on the TwitSuicide dataset. For DEPTWEET, IdenDep, and SDCNL datasets, the best performances of our model are 6.03\%, 5.40\%, 4.40\%, and 2.88\%, 4.50\%, 4.37\% higher than the best-performing baseline model on the macro and weighted F1 scores, respectively.

\subsection{Audio Representation Analysis}

\begin{table}[t]
\footnotesize
\centering
\caption{Samples for the TwitSuicide audio spectrogram analysis. Each sample has been masked to avoid a reverse search of each post. SI: Safe to Ignore; SC: Seriously Concerning.}
\begin{tabularx}{\columnwidth}{
    >{\RaggedRight\arraybackslash}b{0.05\columnwidth} |
    >{\centering\arraybackslash}b{0.06\columnwidth} |
    X 
    }
    \toprule  
    \textbf{ID} & \textbf{Class} & \textbf{Text} \\
    \midrule
    \multicolumn{3}{l}{\textbf{0-10 seconds}} \\
    \midrule
    214 & SI & \_USER\_ *** i can't get that link to work \\
    251 & SI & \_USER\_ or, *** anyone from the *** Rookies all-female racers team \\
    355 & SI & *** kill myself.. watching Drag Me To Hell \\
    457 & SI & i'm afraid my ups might be dead *** making ticking noises \\
    495 & SI & It's too early to be awake *** got up 3 1/2 hours ago! *** never wake up before 8. \\
    152 & SC & \_USER\_ *** never wanted to be dead til now... \\
    187 & SC & \_USER\_ thanks...now *** me kill myself \\
    336 & SC & feeling like death *** want to die \\
    407 & SC & I *** die right now no one loves me \\
    573 & SC & *** hate my life sometimes i just want to die \\
    \midrule
    \multicolumn{3}{l}{\textbf{10-25 seconds}} \\
    \midrule


    92 & SI & \_USER\_ Its a story about how success as *** columnist, *** helped create, returned him to alcohol \&amp; suicidal thoughts \\
    216 & SI & \_USER\_ Thx for *your* part ***! Any time you reblog *** I instantly get 3-5x the activity on it I normally do! \\
    359 & SI & Gosh back to *** from *** with a rather large BANG. Really don't want to be here, much rather be sailing *** for lunch \\
    506 & SI & Last paper ***! Lucky I've bought *** or I will be dead. \\
    553 & SI & oh noesss *** is dieing im ganna kill myself!!! *** room with no phone fml. \\
    81 & SC & \_USER\_ ugh. *** that one was awful. suicide always *** gets to me... \\
    160 & SC & \_USER\_ *** so stupid! *** think I still can? I want to kill myself! \\
    416 & SC & I need to go on suicide watch. ***, The Fumble, ***, Jose Mesa, ***, and now this... Where's my razor blade? \\
    417 & SC & I never have any one to talk to *** i hate my self *** kill myself if no one *** say anything to me on *** \\
    589 & SC & So ***. I'm in pain. Sucks. That was *** the point. suicide an option? \\
    
    \bottomrule 
\end{tabularx}
\label{tbl:audioanalysissamplestwitter}
\end{table}

\begin{table}[t]
\footnotesize
\centering
\caption{Samples for the DEPTWEET audio spectrogram analysis. Each sample has been masked to avoid a reverse search of each post. ND: Non-Depressed; SE: Severe.}
\begin{tabularx}{\columnwidth}{
    >{\RaggedRight\arraybackslash}b{0.05\columnwidth}  | 
    >{\centering\arraybackslash}b{0.06\columnwidth}  | 
    X
    }
    \toprule  
    \textbf{ID} & \textbf{Class} & \textbf{Text} \\
    \midrule
    \multicolumn{3}{l}{\textbf{0-10 seconds}} \\
    \midrule


    665	& ND &	*** miss my sc I’m so depressed without it ☹️☹️	\\
    965	& ND &	\_USER\_ \_USER\_ \_USER\_ Frustrated *** fan hai \_URL\_	\\
    1142	& ND &	Me checking *** I hate *** Continues to check *** and then gets depressed	\\
    2054	& ND &	*** so exhausted *** fighting to stay up until 8pm	\\
    3107	& ND &	Do you feel frustrated *** on the simplest things? \_USER\_ ...	\\
    3310	& ND &	*** teacher is so tired of *** shit	\\
    1085	& SE &	\_USER\_ sh000t me it would hurt less ***	\\
    2442	& SE &	*** so lonely. *** going to hurt someone . \#depression \_USER\_	\\
    4389	& SE &	We *** the shit country. *** so depressed. \_URL\_	\\
    4552	& SE &	*** no reason to live. *** I'll just end it . \#depression \_USER\_	\\
    
    \midrule
    \multicolumn{3}{l}{\textbf{10-25 seconds}} \\
    \midrule

    159	& ND &	\_USER\_ Neither. *** people who *** clinically depressed are going to be so regardless of their worldview. IMHO	\\
    1849	& ND &	ofc *** days off im taking care of my newphew ... im so tired i work every weekday *** takes forever to get home *** on my days off i babysit... i don’t even get paid. im exhausted	\\
    2896	& ND &	Football: *** revive World Cup hopes, *** frustrated by *** \_URL\_	\\
    3382	& ND &	*** sad of getting old it made us restless... *** so MAD i'm getting old it makes me reckless!!!	\\
    4158	& ND &	I've *** my toenails off and split the nail bed - the pain has progressed over *** days to absolutely excruciating - so bad *** struggling to even walk. This week is going amazing	\\
    230	& SE &	*** first guest: Me. *** self-sabotage and self-destruction.	\\
    1686	& SE &	\_USER\_ Man, September was so hard *** watched my gma pass away, *** so much other stuff went wrong. I been depressed asf	\\
    2807	& SE &	\_USER\_ \_USER\_ \_USER\_ \_USER\_ \_USER\_ I personally can't *** 3 or4 died *** from either trauma or anxiety and *** those who took their own lives because of what happened	\\
    2919	& SE &	*** get the hell out. so I'll just end it . \#depression \_USER\_	\\
    4838	& SE &	*** thinking about suicide more and more *** I don’t want to. I don’t want *** that trauma on my kid. But it’s hard… *** suffering from depression *** 15 years… it’s a daily battle… I’m tired	\\
    \bottomrule 
\end{tabularx}
\label{tbl:audioanalysissamplesdeptweet}
\end{table}

\begin{table}[t]
\footnotesize
\centering
\caption{Samples for the IdenDep audio spectrogram analysis. Each sample has been masked to avoid a reverse search of each post. NDE: Non-Depressive; DE: Depressive.}
\begin{tabularx}{\columnwidth}{
    >{\RaggedRight\arraybackslash}b{0.05\columnwidth}  | 
    >{\centering\arraybackslash}b{0.06\columnwidth}  | 
    X
    }
    \toprule  
    \textbf{ID} & \textbf{Class} & \textbf{Text} \\
    \midrule
    \multicolumn{3}{l}{\textbf{0-10 seconds}} \\
    \midrule


    1324	& NDE &	*** Sympathy gift ideas + ***	\\
    1427	& NDE &	*** friend vlog	\\
    1701	& NDE &	does anyone want to hear the story about *** 'beef' between ***	\\
    1728	& NDE &	TRUST by " THE HIPSTERS " (ft. *** and ***)	\\
    1754	& NDE &	How to have a strong family *** products or services have helped *** family stay strong together?	\\
    172	& DE &	It's *** easier to fall back in than to fight *** it	\\
    655	& DE &	*** friends are throwing a LAN party *** I wasn't invited. *** only one who didn't get an invitation.	\\
    904	& DE &	*** feel bitter about everything. *** bitter about being bitter.	\\
    934	& DE &	I'm sad I feel sad.  *** I feel something.	\\
    1235	& DE &	... I just want to crawl in a whole and cry ***	\\
    
    \midrule
    \multicolumn{3}{l}{\textbf{10-25 seconds}} \\
    \midrule

    1377	& NDE &	Having friends *** opposite sex *** in a relationship \_URL\_	\\
    1470	& NDE &	*** introduce my girlfriend [18F] to my family My girlfriend lives *** I live in *** *** introduce her to my mom but I don't know how	\\
    1589	& NDE &	365 New Ways To Hug Your Love *** discover and post videos or pictures of New Ways To Hug in the new subreddit ***	\\
    1616	& NDE &	With family being a main interest in your lives, what *** would you purchase *** to help the family to grow?	\\
    1824	& NDE &	What Would You Do? Would you move away from your family *** to somewhere far where your kids would have a better education *** provide for your family better, like buying a house; *** moving from *** to the *** or ***?	\\
    132	& DE &	Anyone else feel like everyone hates them? *** paranoia? *** the dark cloud over my head just gives off a shitty vibe *** people think I don't like them and vice Versa.	\\
    505	& DE &	That feeling when you hate who you *** but can't *** change because you are so used to being like this for *** years. *** a shitty person. The thought of change seems impossible *** at this point.	\\
    605	& DE &	Fuck me When you’re *** a piece of shit *** look at other girls and lie to you, while lying *** next to you. I’ll never be enough. Ever. For anyone. *** want to ducking die.	\\
    1027	& DE &	Addicted to depression *** when I feel like *** self-loathing and depressive *** becoming less, I feel shit *** don't feel depressed anymore. *** I want it to go away *** part of me wants to stay depressive and feel suicidal.	\\
    1154	& DE &	Is it depression *** don't want to build memories anymore? *** I get really nostalgic. *** I don't want to get too attached to people *** just end up hurting in the future.	\\
    
    \bottomrule 
\end{tabularx}
\label{tbl:audioanalysissamplesidendep}
\end{table}

\begin{figure}[t]
    \centering
     \includegraphics[scale=0.48]{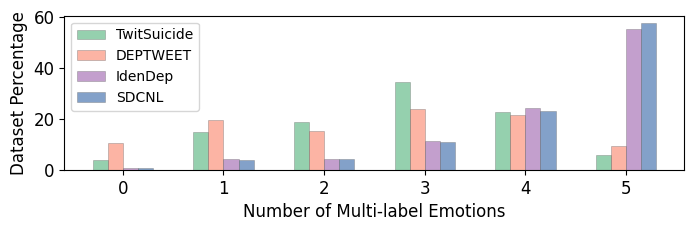}
     \caption{Distribution of multi-label emotion class labels.}
     \label{fig:MultiLabelEmo}
    \hfill
\end{figure}

To ensure the feasibility of our audio modality for mental health detection, we give an illustrative visualisation of the audio embeddings, which are generated by input text and learned via the Audio Spectrogram Transformer (AST). We conduct Principal Component Analysis (PCA) to visualise the acquired audio embeddings and their corresponding mental health class labels. In order to emphasise the distinguishability of the embeddings,  we select samples from both the least and most concerning labels in each dataset, as shown in Figure \ref{fig:AudioAnalysis}.
For each dataset, we group all the generated audio based on durations of 0-to-10-second and 10-to-25-second length\footnote{Note that most generated audios are less than 25 seconds.}. 
For each of these two audio groups, we generated the corresponding spectrograms and randomly selected ten audio samples for each group to visualise the first two principal components after performing PCA.
In Figure \ref{fig:AudioAnalysis}, we annotate each sample with the post ID and the audio duration for detailed comparison.
Tables \ref{tbl:audioanalysissamplestwitter} to \ref{tbl:audioanalysissamplesidendep} contains de-identified and masked post contents for TwitSuicide, DEPTWEET, and IdenDep, respectively. SDCNL samples may be found in the Supplementary Material.

The visualisation shows that our audio embeddings can show a noticeable separation between mental health classes for all four datasets. In datasets derived from Twitter, shorter audio samples display more pronounced distinctions between the most and least concerning classes, whereas, in datasets from Reddit, this separation becomes more evident in longer audio segments. We assume that this is primarily due to Twitter posts being generally shorter in length, whereas Reddit posts tend to be longer. 

\subsection{Effectiveness of Multimodal Multi-Teachers}
\label{sec:AblationStudy}

\begin{table}[t]
\footnotesize
\centering
\caption{Ablation study using different combinations of teacher modalities. Class abbreviation definitions may be found in the Figure \ref{fig:ClassDist} caption. \textbf{Bold} face indicates best score while second best are \underline{underlined}. A \checkmark indicates the addition of the emotion (Emo) and/or the audio (Aud) teacher/s. \colorbox{LightCyan}{Highlighted} rows show the best setup.}
\begin{tabularx}{\columnwidth}{X X X | X X X | X X X X }
    \toprule  
        \textbf{Text$^\ddagger$}
        & \textbf{Emo}
        & \textbf{Aud}     
        &\multicolumn{3}{c|}{\textbf{Overall Performance}} 
        &\multicolumn{4}{c}{\textbf{Breakdown F1 Scores}}\\
        
    \toprule
    \multicolumn{3}{c |}{\textbf{TwitSuicide} } 
        & \textbf{Acc} 
        & \textbf{F1m} 
        & \textbf{F1w} 
        & \textbf{(SC)} 
        & \textbf{(PC)} 
        & \textbf{(SI)} \\
    \hline
    \checkmark
        & $\times$
        & $\times$
        & 58.94
        & 47.50
        & 56.13
        & 16.13
        & 56.55
        & 69.81
        \\
    \rowcolor{LightCyan}\checkmark
        & \checkmark
        & $\times$
        & \textbf{65.76}
        & \textbf{61.96}
        & \textbf{65.46}
        & \underline{49.72}
        & \textbf{62.34}
        & \textbf{73.81}\\
    \checkmark
        & $\times$
        & \checkmark
        & \underline{63.79}
        & 58.94	
        & \underline{63.02}	
        & 44.30	
        & \underline{61.76}	
        & \underline{70.74}\\
    \checkmark
        & \checkmark
        & \checkmark
        & 61.21
        & \underline{59.64}
        & 61.23
        & \textbf{54.17}
        & 59.50
        & 65.26\\
        
    \toprule        
    \multicolumn{3}{c |}{\textbf{DEPTWEET} }   
        & \textbf{Acc}     
        & \textbf{F1m}     
        & \textbf{F1w}        
        & \textbf{(SE)}        
        & \textbf{(MO)}         
        & \textbf{(MI)}              
        & \textbf{(ND)}  \\
    \hline   
        \checkmark
        & $\times$
        & $\times$
        & \textbf{84.52} 
        & 44.33
        & 82.40 
        & \underline{40.00}
        & 13.11 
        & 31.28 
        & \underline{92.90}  \\
    \rowcolor{LightCyan}\checkmark
        & \checkmark
        & $\times$        
        & \underline{84.03} 
        & \textbf{46.43}
        & \textbf{83.09}
        & \textbf{41.38}  
        & 12.31  
        & \textbf{39.07} 
        & \textbf{92.95}\\
    \checkmark
        & $\times$
        & \checkmark
        & 83.06 
        & 36.00
        & 81.69  
        & 00.00 
        & 15.38
        & \underline{36.26} 
        & 92.33 \\
    \checkmark
        & \checkmark
        & \checkmark
        & 82.77 
        & \underline{46.20}
        & \underline{82.61} 
        & 26.09 
        & \textbf{34.34} 
        & 32.32 
        & 92.04 \\
    
    \toprule    
    \multicolumn{3}{c |}{\textbf{IdenDep} } 
        & \textbf{Acc} 
        & \textbf{F1m}
        & \textbf{F1w} 
        & \textbf{(DE)} 
        & \textbf{(NDE)} \\
    \hline   
    \checkmark
        & $\times$
        & $\times$
        & 92.32
        & 90.67
        & 92.26
        & 94.60
        & 86.74 \\
    \checkmark
        & \checkmark
        & $\times$
        & 93.86
        & 92.47
        & 93.78
        & 85.71
        & 89.23\\
    \rowcolor{LightCyan}\checkmark
        & $\times$
        & \checkmark
        & \textbf{94.30}
        & \textbf{93.10}
        & \textbf{94.26}
        & \textbf{95.97}
        & \textbf{90.23}\\
    \checkmark
        & \checkmark
        & \checkmark
        & \underline{93.92}
        & \underline{92.58}
        & \underline{93.85}
        & \underline{95.73}
        & \underline{89.43}\\

    \toprule
    \multicolumn{3}{c |}{\textbf{SDCNL} } 
        & \textbf{Acc} 
        & \textbf{F1m}
        & \textbf{F1w} 
        & \textbf{(SUI)} 
        & \textbf{(DEP)} \\
    \hline   
    \checkmark
        & $\times$
        & $\times$
        & \underline{75.20}
        & \underline{75.07}
        & \underline{75.10}
        & 76.85
        & \underline{73.30}\\
    \checkmark
        & \checkmark
        & $\times$
        & 72.82
        & 72.45
        & 72.51
        & 75.65
        & 69.25\\
    \rowcolor{LightCyan}\checkmark
        & $\times$
        & \checkmark
        & \textbf{76.52}
        & \textbf{76.50}
        & \textbf{76.51}
        & \underline{77.12}
        & \textbf{75.88}\\
    \checkmark
        & \checkmark
        & \checkmark
        & \underline{75.20}
        & 74.84
        & 74.90
        & \textbf{77.83}
        & 71.86\\
    \bottomrule
\end{tabularx}
\flushleft
\footnotesize{$^\ddagger$We report results from the best performing pre-trained language model for each dataset (Table \ref{tbl:ablationstudytextteachers}): MentalBERT for DEPTWEET and BERT for the others.}
\label{tbl:ablationstudyteachers}
\end{table}

\begin{table}[t]
\footnotesize
\centering
\caption{Ablation study using different PLMs for the text-based teacher. We report results using the best-performing teacher modality combination in Table \ref{tbl:ablationstudyteachers} and change only the text-based teacher. Class abbreviation definitions may be found in the Figure \ref{fig:ClassDist} caption. \textbf{Bold} face indicates best score while second best are \underline{underlined}.}
\begin{tabularx}{\columnwidth}{
    X |
    >{\centering\arraybackslash}b{0.06\columnwidth}
    >{\centering\arraybackslash}b{0.06\columnwidth}
    >{\centering\arraybackslash}b{0.06\columnwidth} |
    >{\centering\arraybackslash}b{0.06\columnwidth} 
    >{\centering\arraybackslash}b{0.06\columnwidth} 
    >{\centering\arraybackslash}b{0.06\columnwidth} 
    >{\centering\arraybackslash}b{0.06\columnwidth}     
    }
    \toprule  
        &\multicolumn{3}{c|}{\textbf{Overall Performance}} 
        &\multicolumn{4}{c}{\textbf{Breakdown F1 Scores}}\\ 
        
    \toprule
    \textbf{TwitSuicide}
        & \textbf{Acc} 
        & \textbf{F1m} 
        & \textbf{F1w} 
        & \textbf{(SC)} 
        & \textbf{(PC)} 
        & \textbf{(SI)} \\
    \hline
    BERT            & \textbf{65.76} & \underline{61.96} & \textbf{65.46} & 49.72 & \underline{62.34} & \textbf{73.81}\\
    RoBERTa         & 65.15 & 61.67 & 64.70 & 51.19 & 61.43 & \underline{72.40}\\
    MentalBERT      & 63.79 & 61.08 & 63.67 & \underline{51.65} & \textbf{64.08} & 67.52\\
    ClinicalBERT    & \underline{65.30} & \textbf{63.23} & \underline{65.25} & \textbf{56.38} & 62.18 & 71.13\\
        
    \toprule        
    \textbf{DEPTWEET}
        & \textbf{Acc}     
        & \textbf{F1m}     
        & \textbf{F1w}             
        & \textbf{(SE)}           
        & \textbf{(MO)}         
        & \textbf{(MI)}       
        & \textbf{(ND)} \\
    \hline   
    BERT            & \underline{83.84} & \underline{39.80} & \textbf{83.84} & 0.00 & \textbf{22.22}  & \textbf{44.05} & \underline{92.93}\\
    RoBERTa         & 82.67 & 35.99 & 81.80 & 0.00 & \underline{14.29} & 37.23 & 92.44\\
    MentalBERT      & \textbf{84.03} & \textbf{46.43} & \underline{83.09} & \textbf{41.38} & 12.31 & \underline{39.07} & \textbf{92.95}\\
    ClinicalBERT    & 83.54 & 32.29 & 80.71 & 0.00 & 0.00 & 37.11 & 92.05\\
    
    \toprule    
    \textbf{IdenDep}
        & \textbf{Acc} 
        & \textbf{F1m}
        & \textbf{F1w} 
        & \textbf{(DE)} 
        & \textbf{(NDE)} \\
    \hline   
    BERT            & \textbf{94.30} & \textbf{93.10} & \textbf{94.26} & \textbf{95.97} & \textbf{90.23} \\
    RoBERTa         & 94.13 & 92.89 & 94.09 & \underline{95.86} & 89.93 \\
    MentalBERT      & 93.21 & 91.71 & 93.14 & 95.24 & 88.17 \\
    ClinicalBERT    & \underline{94.24} & \underline{92.95} & \underline{94.17} & \textbf{95.97} & \underline{89.94} \\

    \toprule
    \textbf{SDCNL}
        & \textbf{Acc} 
        & \textbf{F1m}
        & \textbf{F1w} 
        & \textbf{(SUI)} 
        & \textbf{(DEP)} \\
    \hline   
    BERT            & \textbf{76.52} & \textbf{76.50} & \textbf{76.51} & \textbf{77.12} & \textbf{75.88} \\
    RoBERTa         & 75.20 & 75.03 & 75.07 & \underline{77.07} & 72.99 \\
    MentalBERT      & 73.61 & 73.61 & 73.61 & 73.40 & 73.82 \\
    ClinicalBERT    & \underline{75.46} & \underline{75.46} & \underline{75.45} & 75.07 & \underline{75.84} \\
    \bottomrule
\end{tabularx}
\flushleft
\label{tbl:ablationstudytextteachers}
\end{table}

To examine the efficacy of each teacher modality and their combinations, we evaluate and explore the performance by adding an extra emotion-based teacher, an audio-based teacher, or both alongside the text-based teacher. The results are presented in Table \ref{tbl:ablationstudyteachers}.

In general, the multi-teacher structure outperforms the use of a singular text-based teacher, although the effectiveness of different modalities varies across datasets. For the Twitter-based datasets,
applying the emotion-based and text-based teachers together achieves the best results. In contrast, for Reddit-based datasets,
the audio- and text-based teachers have better performances. This difference may be attributed to the longer posts in Reddit-based datasets, resulting in longer audio (Table \ref{tbl:datastatistics}) that contains more acoustic information beneficial for the audio-based teacher.
Moreover, due to the lengthier nature of posts in Reddit-based datasets, more SenticNet lexicon tokens are likely to be matched compared to Twitter-based datasets. This results in a higher number of generated emotion labels during the learning process of the multi-label emotion-based teacher. In Figure \ref{fig:MultiLabelEmo}, we compare the count of generated multi-label emotion classes utilised to train the emotion-based teacher across all four datasets. It is evident that a greater proportion of posts in Reddit-based datasets match all seven emotion labels (shown in Section \ref{sub:emoteacher}) compared to Twitter-based datasets. 
This potential increase in the number of matching emotion labels may present challenges in distinguishing between different emotions during the training of the emotion-based teacher, potentially impacting downstream mental health classification, especially for ambiguous classes such as \textit{Suicide} (SUI) and \textit{Depression} (DEP) in the SDCNL dataset.

We can conclude that using multimodality teachers generally helps detect mental health, and these findings also suggest varying effectiveness of different modalities across datasets with distinct characteristics, offering valuable insights into selecting suitable modalities for improved performance in future scenarios.

\subsection{Impact of Text-based Teachers}
\label{sec:ablationlanguageteachers}

We compare the effectiveness of various PLMs for the text-based teacher. Table \ref{tbl:ablationstudytextteachers} shows that BERT produces the best weighted F1 across all datasets.
However, the domain-specific PLMs perform better for the more concerning classes in a multi-class setup. ClinicalBERT outperforms BERT by 6.66\% for \textit{Strongly Concerning} (SC) in the TwitSuicide dataset, while MentalBERT achieves 41.38\% for \textit{Severe} (SE) in the DEPTWEET dataset, surpassing the other language models which failed to recognise it. To ensure optimal performance, we specifically employ MentalBERT for DEPTWEET, while BERT is used for the other datasets. Nonetheless, the overall performance of the text-based teacher is not significantly impacted by the choice of PLMs. 
More performance enhancement stems from the inclusion of different modalities, as discussed in the previous sections.

\subsection{Impact of Student Model Inputs}

\begin{table}[t]
\footnotesize
\centering
\caption{Ablation study using different combinations of input modalities to the student model. \textbf{Bold} face indicates best score while second best are \underline{underlined}. A \checkmark indicates the addition of the emotion-based (Emo) and/or the audio-based (Aud) input features. \colorbox{LightCyan}{Highlighted} rows show our proposed student setup. VT: randomly initialised vanilla transformer.}
\begin{tabularx}{\columnwidth}{
    X X X | X X X | X X X X  
    }
    \toprule  
        \textbf{Text}
        & \textbf{Emo}
        & \textbf{Aud}        
        & \multicolumn{3}{c|}{\textbf{Overall Performance}} 
        & \multicolumn{4}{c}{\textbf{Breakdown F1 Scores}} 
        \\
        
    \toprule
    \multicolumn{3}{c |}{\textbf{TwitSuicide} } 
        & \textbf{Acc} 
        & \textbf{F1m} 
        & \textbf{F1w} 
        & \textbf{(SC)} 
        & \textbf{(PC)} 
        & \textbf{(SI)} \\
    \hline
    BERT & \checkmark & \checkmark              & \underline{51.67}	& \underline{45.43}	& \underline{50.46}	& 26.95	& \underline{52.00}	& 57.34 \\
    BERT & \checkmark & $\times$                  & 50.91	& 44.24	& 48.58	& \underline{29.58}	& 41.42	& \underline{61.71} \\
    BERT & $\times$ & \checkmark                  & 48.64	& 33.87	& 43.31	& 0.00	& 41.07	& 60.55 \\
    \rowcolor{LightCyan}BERT & $\times$ & $\times$  & \textbf{65.76}	& \textbf{61.96}	& \textbf{65.46}	& \textbf{49.72}	& \textbf{62.34}	& \textbf{73.81} \\
    VT & $\times$ & $\times$                        & 46.36	& 32.72	& 41.77	& 0.00	& 41.09	& 57.06 \\
        
    \toprule        
    \multicolumn{3}{c |}{\textbf{DEPTWEET} }   
        & \textbf{Acc}     
        & \textbf{F1m}     
        & \textbf{F1w}            
        & \textbf{(SE)}          
        & \textbf{(MO)}         
        & \textbf{(MI)}        
        & \textbf{(ND)}  \\
    \hline   
    BERT & \checkmark & \checkmark              & \textbf{85.49} & \underline{34.75} & \underline{81.24} & 0.00 & \textbf{20.59} & 25.86 & 92.54 \\
    BERT & \checkmark & $\times$                  & 83.84 & 32.15 & 80.81 & 0.00 & 0.00  & \underline{36.36} & \underline{92.55}\\
    BERT & $\times$ & \checkmark                  & \underline{84.23} & 22.86 & 77.01 & 0.00 & 0.00 & 0.00 & 91.44  \\
    \rowcolor{LightCyan}BERT & $\times$ & $\times$  & 84.03 & \textbf{46.43} & \textbf{83.09} & \textbf{41.38} & \underline{12.31} & \textbf{39.07} & \textbf{92.95} \\
    VT & $\times$ & $\times$                        & \underline{84.23} & 22.86 & 77.01 & 0.00 & 0.00 & 0.00 & 91.44  \\
    
    \toprule    
    \multicolumn{3}{c |}{\textbf{IdenDep} } 
        & \textbf{Acc} 
        & \textbf{F1m}
        & \textbf{F1w} 
        & \textbf{(DE)} 
        & \textbf{(NDE)} \\
    \hline   
    BERT & \checkmark & \checkmark              & 91.58 & \underline{89.99} & 91.60 & 93.98 & \underline{86.00} \\
    BERT & \checkmark & $\times$                  & 91.09 & 89.21 & 91.03 & 93.72 & 84.70 \\
    BERT & $\times$ & \checkmark                  & \underline{91.75} & 89.85 & \underline{91.62} & \underline{94.23} & 85.47 \\
    \rowcolor{LightCyan}BERT & $\times$ & $\times$  & \textbf{94.30} & \textbf{93.10} & \textbf{94.26} & \textbf{95.97} & \textbf{90.23} \\
    VT & $\times$ & $\times$                        & 75.77 & 61.25 & 70.85 & 84.97 & 37.54 \\

    \toprule
    \multicolumn{3}{c |}{\textbf{SDCNL} } 
        & \textbf{Acc} 
        & \textbf{F1m}
        & \textbf{F1w} 
        & \textbf{(SUI)} 
        & \textbf{(DEP)} \\
    \hline   
    BERT & \checkmark & \checkmark              & 67.55 & 67.54 & 67.55 & 67.89 & 67.20 \\
    BERT & \checkmark & $\times$                  & 67.81 & 67.80 & 67.80 & 67.38 & 68.23 \\
    BERT & $\times$ & \checkmark                  & \underline{68.34} & \underline{68.32} & \underline{68.31} & 67.57 & \underline{69.07} \\
    \rowcolor{LightCyan}BERT & $\times$ & $\times$  & \textbf{76.52} & \textbf{76.50} & \textbf{76.51} & \textbf{77.12} & \textbf{75.88} \\
    VT & $\times$ & $\times$                        & 67.02 & 66.94 & 66.97 & \underline{68.51} & 65.37 \\
    \bottomrule
\end{tabularx}
\label{tbl:ablationstudystudent}
\end{table}

\begin{figure}[t]
    \captionsetup[sub]{justification=centering}
    \centering

    \begin{subfigure}[t]{0.48\columnwidth}
         \centering
         \includegraphics[scale=0.25]{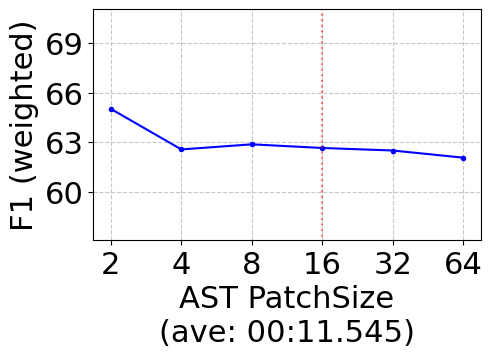}
         \caption{TwitSuicide}
         \label{fig:AudioTwitter}
    \end{subfigure}    
    \begin{subfigure}[t]{0.48\columnwidth}
         \centering
         \includegraphics[scale=0.25]{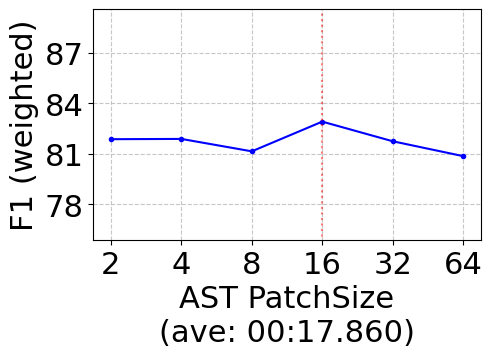}
         \caption{DEPTWEET}
         \label{fig:AudioDEPTWEET}
    \end{subfigure}  
    \begin{subfigure}[t]{0.48\columnwidth}
         \centering
         \includegraphics[scale=0.25]{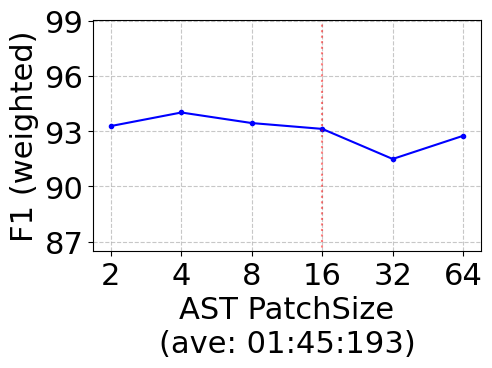}
         \caption{IdenDep}
         \label{fig:AudioIdenDep}
    \end{subfigure}  
    \begin{subfigure}[t]{0.48\columnwidth}
         \centering
         \includegraphics[scale=0.25]{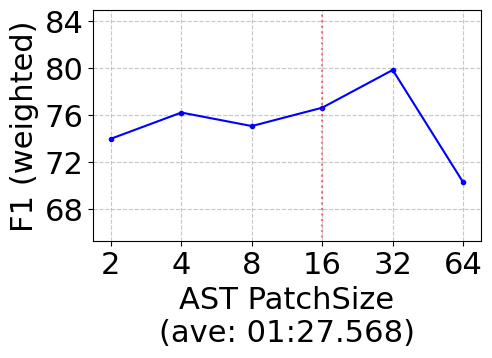}
         \caption{SDCNL}
         \label{fig:AudioSDCNL}
    \end{subfigure}  
    \caption{Parameter study for the audio-based teacher model. Ave: average audio duration for the dataset.}
    \label{fig:HypTest}
\end{figure}

We examine different combinations of multimodal inputs for the student model in Table \ref{tbl:ablationstudystudent} in order to explore the optimal input for a knowledge distillation for the student model. We concatenate emotion embeddings, audio embeddings, or both with text embeddings from pre-trained BERT and then pass them to the transformer layer after a linear layer projection. We also test a randomly initialised vanilla transformer compared to the pre-trained BERT. The results indicate that unimodal textual post inputs outperform the concatenation of multimodal inputs for the student model. Moreover, pre-trained BERT yields better results than the randomly initialised vanilla transformer across all datasets. These outcomes underscore the effectiveness of multi-aspect knowledge acquired from the multi-teachers, efficiently guiding the student to achieve robust performance with only textual inputs.

\subsection{Hyperparameter Testing}
\label{sec:Hyperparameter Testing}

We further investigate the different patch size values for the audio-based teacher model while maintaining a consistent setup for the student model. Figure \ref{fig:HypTest} shows each dataset's weighted F1 score for each patch size value. TwitSuicide, DEPTWEET, and IdenDep datasets show a relatively stable performance between 2 to 64 patch sizes; however, for SDCNL, performance improvement may be achieved using a patch size of 32. This may be due to the higher variance in audio duration of the outliers compared to the other three datasets (Figure \ref{fig:AudioBoxOutlier}). Despite being shorter on average length and duration than the other Reddit-based dataset, SDCNL has some longer audio samples, which may have benefited from a patch size 32. However, a sharp decline in performance could be expected when the patch size is increased to 64.

\section{Conclusion}
In conclusion, our study introduces a multimodal multi-teacher knowledge distillation model, 3M-Health, designed for mental health detection and presents a comprehensive exploration. Our experiments demonstrate that the multimodal approach outperforms unimodal counterparts, with the choice of modalities influencing performance across diverse datasets. Notably, the incorporation of audio-based information proves particularly beneficial for social media post-based mental health detection for Reddit-based datasets, emphasising the importance of modality selection based on the nature of the data. Overall, our work contributes valuable insights into the nuanced dynamics of multimodal knowledge distillation for mental health detection, offering a promising avenue for future research in this critical domain.

\begin{acks}
This study was supported by funding from the Google Award for Inclusion Research Program
(G222897).
\end{acks}


\bibliographystyle{ACM-Reference-Format}
\balance
\bibliography{sidsm}

\end{document}


\title{3M-Health: Multimodal Multi-Teacher Knowledge Distillation for Mental Health Detection}
\titlenote{\textcolor{red}{Warning: This paper contains examples that show suicide ideation and depression.}}
\subtitle{Supplementary Material}

\author{Rina Carines Cabral}
\affiliation{%
    \institution{The University of Sydney}
    \city{Sydney}
    \state{NSW}
    \postcode{2006}
    \country{Australia}
}
\email{rcab5321@uni.sydney.edu.au}
\orcid{0000-0003-3076-0521}

\author{Siwen Luo}
\affiliation{
    \institution{The University of Western Australia}
    \city{Perth}
    \state{WA}
    \postcode{6009}
    \country{Australia}
}
\email{siwen.luo@uwa.edu.au}
\orcid{0000-0003-0480-1991}

\author{Josiah Poon}
\affiliation{
    \institution{The University of Sydney}
    \city{Sydney}
    \state{NSW}
    \postcode{2006}
    \country{Australia}
}
\email{josiah.poon@sydney.edu.au}
\orcid{0000-0003-3371-8628}

\author{Soyeon Caren Han}
\authornote{Corresponding author.}
\affiliation{
    \institution{The University of Melbourne}
    \city{Melbourne}
    \state{VIC}
    \postcode{3052}
    \country{Australia}
}
\email{caren.han@unimelb.edu.au}
\orcid{0000-0002-1948-6819}

\renewcommand{\shortauthors}{Rina Carines Cabral, Siwen Luo, Josiah Poon, and Soyeon Caren Han}

\maketitle

\appendix

\section{Hyperparameter Search}
\label{adx:Hyperparameter Search}

\setcounter{figure}{0}
\setcounter{table}{0}
\renewcommand{\thefigure}{A\arabic{figure}}
\renewcommand{\thetable}{A\arabic{table}}

We tune each individual model for our experiments using Optuna\footnote{https://optuna.org/} for 50 trials and optimise the weighted F1 score. Final model construction is done separately from model tuning using a 90:10 validation split. Table \ref{tbl:hyperparametersearch} details the parameters and the search space used. Table \ref{tbl:teacherhyperparameters} and \ref{tbl:studenthyperparameters} enumerates the best-found hyperparameters for the final teacher models and student models, respectively.

\begin{table}[H]
\footnotesize
\centering
\caption{Hyperparameter Search Space}
\begin{tabularx}{\columnwidth}{
    >{\RaggedRight\arraybackslash}p{0.25\columnwidth} |
    X
    }
    \toprule
    \textbf{Parameter}
        & \textbf{Search Space}
        \\
    \midrule
    \multicolumn{2}{l}{\textbf{Text-based Teachers}} \\
    \midrule
    Dropout &  \{0.01, 0.05, 0.1, 0.5\} \\
    Hidden Layers &  \{2, 4, 6, 8, 10, 12\} \\
    Attention Heads &  \{2, 4, 6, 8, 12\} \\
    Learning Rate & \{1e-04, 1e-05, 2e-05, 3e-05, 4e-05, 5e-05\} \\
    Weight Decay & \{0, 0.01, 0.1\} \\
    Epochs & [2-5] \\
    \midrule
    \multicolumn{2}{l}{\textbf{Emotion-based Teachers}} \\
    \midrule
    Dropout & \{0.01, 0.05, 0.1, 0.5\} \\
    Hidden Layers & [2-5] \\
    Hidden Dimension & \{100, 200, 300, 400, 500\} \\
    Learning Rate & \{1e-03, 1e-04, 1e-05\} \\
    Weight Decay & \{0, 0.01, 0.1\} \\
    \midrule
    \multicolumn{2}{l}{\textbf{Audio-based Teachers}} \\
    \midrule  
    Dropout &  \{0.01, 0.05, 0.1, 0.5\} \\
    Hidden Layers &  \{2, 4, 6, 8, 10, 12\} \\
    Attention Heads &  \{2, 4, 6, 8, 12\} \\
    Learning Rate & \{1e-03, 1e-04, 1e-05, 5e-05\} \\
    Scheduler Patience & [2-5] \\
    Scheduler Factor & \{0.1, 0.5\} \\
    \midrule
    \multicolumn{2}{l}{\textbf{Student}} \\
    \midrule
    Dropout & \{0.01, 0.05, 0.1, 0.5\} \\
    Learning Rate & \{1e-04, 1e-05, 2e-05, 3e-05, 4e-05, 5e-05\} \\
    Weight Decay & \{0, 0.01, 0.1\} \\
    Hidden Layers & \{2, 4, 6, 8, 10, 12\} \\
    Attention Heads & \{2, 4, 6, 8, 12\} \\
    Activation & \{``relu”, ``gelu”\} \\
    Epochs & [3-5] \\    
    \bottomrule
\end{tabularx}
\label{tbl:hyperparametersearch}
\end{table}

\begin{table}[H]
\footnotesize
\centering
\caption{Best found teacher hyperparameters.}
\begin{tabularx}{\columnwidth}{
    X |
    >{\centering\arraybackslash}p{0.14\columnwidth} 
    >{\centering\arraybackslash}p{0.16\columnwidth} 
    >{\centering\arraybackslash}p{0.10\columnwidth} 
    >{\centering\arraybackslash}p{0.10\columnwidth}
    }
    \toprule
        \textbf{Parameters}
        & \textbf{TwitSuicide} 
        & \textbf{DEPTWEET}
        & \textbf{IdenDep} 
        & \textbf{SDCNL} 
        \\
    \midrule    
    \multicolumn{5}{l}{\textbf{Text-based Teachers}} \\
    \midrule
    Language Model              
        & BERT 
        & MentalBERT 
        & BERT & BERT \\
    Dropout         
        & 0.01 
        & 0.01
        & 0.05 & 0.01 \\
    Weight Decay    
        & 0 
        & 0
        & 0 & 0 \\
    Learning Rate   
        & 4e-05 
        & 4e-05
        & 4e-05 & 5e-05 \\
    Epochs          
        & 4 
        & 3
        & 5 & 4 \\
    Hidden Layers   
        & 10 
        & 8
        & 2 & 4 \\
    Attention Heads 
        & 8 
        & 8 & 6 & 12\\
    Batch Size      
        & 128 
        & 64
        & 64 & 64 \\
    \midrule
    \multicolumn{5}{l}{\textbf{Emotion-based Teachers}} \\
    \midrule
    Dropout         
        & 0.1 
        & 0.1
        & 0.05 & 0.01 \\
    Weight Decay    
        & 0 
        & 0.01 
        & 0.1 & 0 \\
    Learning Rate   
        & 1e-05 
        & 1e-04 
        & 1e-04 & 1e-05 \\
    Hidden Layers   
        & 2 
        & 4 
        & 3 & 5 \\
    Hidden Dim      
        & 400 
        & 400 
        & 100 & 100 \\
    Batch Size      
        & 128 
        & 64 
        & 64 & 64 \\
    \midrule
    \multicolumn{5}{l}{\textbf{Audio-based Teachers}} \\
    \midrule
    Dropout         
        & 0.1 
        & 0.1
        & 0.01 & 0.01 \\
    Learning Rate   
        & 5e-05 
        & 1e-05
        & 1e-05 & 5e-05 \\
    Hidden Layers   
        & 8 
        & 6
        & 6 & 8 \\
    Attention Heads      
        & 4 
        & 6
        & 8 & 8 \\
    Scheduler Patience   
        & 4 
        & 4
        & 3 & 4 \\
    Scheduler Factor     
        & 0.5 
        & 0.5
        & 0.5 & 0.1 \\
    Batch Size      
        & 32 
        & 32
        & 32 & 32 \\
    \bottomrule
\end{tabularx}
\label{tbl:teacherhyperparameters}
\end{table}

\begin{table}[H]
\footnotesize
\centering
\caption{Best found student hyperparameters for different combinations of teacher modalities.}
\begin{tabularx}{\columnwidth}{
    X |
    >{\centering\arraybackslash}p{0.14\columnwidth} 
    >{\centering\arraybackslash}p{0.16\columnwidth} 
    >{\centering\arraybackslash}p{0.10\columnwidth} 
    >{\centering\arraybackslash}p{0.10\columnwidth}
    }
    \toprule
        \textbf{Parameters}
        & \textbf{TwitSuicide} 
        & \textbf{DEPTWEET}
        & \textbf{IdenDep} & \textbf{SDCNL} 
        \\
    \midrule    
    \multicolumn{5}{l}{\textbf{Text \checkmark Emo \checkmark Aud \checkmark}} \\
    \midrule
    Dropout         
        & 0.05 
        & 0.01
        & 0.1 & 0.05 \\
    Learning Rate   
        & 1e-04 
        & 3e-05
        & 5e-05 & 5e-05 \\
    Weight Decay    
        & 0 
        & 0
        & 0 & 0.01 \\
    Hidden Layers   
        & 10 
        & 6
        & 10 & 12 \\
    Attention Heads      
        & 3 
        & 12
        & 8 & 12 \\
    Activation      
        & gelu 
        & gelu
        & gelu & gelu \\
    Epochs          
        & 3 
        & 5
        & 3 & 3 \\    
    
    \midrule    
    
    \multicolumn{5}{l}{\textbf{Text \checkmark Emo \checkmark Aud $\times$}} \\
    \midrule
    Dropout         
        & 0.1 
        & 0.01
        & 0.1 & 0.05 \\
    Learning Rate   
        & 1e-04 
        & 3e-05
        & 5e-05 & 4e-05 \\
    Weight Decay    
        & 0.01 
        & 0
        & 0 & 0 \\
    Hidden Layers   
        & 10 
        & 10
        & 10 & 12 \\
    Attention Heads      
        & 12 
        & 12
        & 6 & 3 \\
    Activation      
        & gelu  
        & gelu
        & gelu & gelu \\
    Epochs          
        & 4 
        & 5
        & 3 & 5 \\
    
    \midrule    
    
    \multicolumn{5}{l}{\textbf{Text \checkmark Emo $\times$ Aud \checkmark}} \\
    \midrule
    Dropout         
        & 0.05 
        & 0.05
        & 0.05 & 0.05 \\
    Learning Rate   
        & 1e-04 
        & 1e-04
        & 4e-05 & 5e-05 \\
    Weight Decay    
        & 0 
        & 0
        & 0 & 0 \\
    Hidden Layers   
        & 12 
        & 4
        & 12 & 12 \\
    Attention Heads      
        & 4 
        & 8
        & 12 & 12 \\
    Activation      
        & gelu 
        & relu
        & gelu & gelu \\
    Epochs          
        & 4 
        & 3
        & 4 & 5 \\
    
    \midrule    
    
    \multicolumn{5}{l}{\textbf{Text \checkmark Emo $\times$ Aud $\times$}} \\
    \midrule
    Dropout         
        & 0.1 
        & 0.01
        & 0.05 & 0.05 \\
    Learning Rate   
        & 1e-04 
        & 4e-05
        & 1e-4 & 4e-05 \\
    Weight Decay    
        & 0 
        & 0
        & 0 & 0 \\
    Hidden Layers   
        & 12 
        & 12
        & 4 & 10 \\
    Attention Heads      
        & 8 
        & 6
        & 4 & 12 \\
    Activation      
        & relu 
        & gelu
        & gelu & relu \\
    Epochs          
        & 3 
        & 4
        & 5 & 4 \\

    \bottomrule
    
\end{tabularx}
\label{tbl:studenthyperparameters}
\end{table}

\section{Audio Analysis Samples}
\label{adx:audiosamples}

\setcounter{figure}{0}
\setcounter{table}{0}
\renewcommand{\thefigure}{B\arabic{figure}}
\renewcommand{\thetable}{B\arabic{table}}

\textcolor{red}{\textbf{Content Warning}: The following section shows samples of posts indicative of suicide and depression, which may be triggering for some people.}

We provide the textual posts associated with each sample presented in Section 5.2 Figure 5 to refer to for detailed comparison and analysis. 
Table \ref{tbl:audioanalysissamplessdcnl} enumerates the post ID, class, and text for the samples from the SDCNL dataset. TwitSuicide, DEPTWEET and IdenDep samples are found in Tables 5 to 7 of the main manuscript.
Following the proposed ethical protocols on social media research of \cite{benton-2017-ethical}, usernames and links are masked with special tokens \textit{\_USER\_} and \textit{\_URL\_} to protect the identity and privacy of each author. For instance, ``@myusername this is the link http://urlsamp.le” is masked as ``\_USER\_ this is the link \_URL\_”. Moreover, we mask parts of the text with “***” to prevent possible reverse searches.

\begin{table}[H]
\footnotesize
\centering
\caption{Samples for the SDCNL audio spectrogram analysis. Each sample has been masked to avoid a reverse search of each post. DEP: Depression; SUI: Suicide}
\begin{tabularx}{\columnwidth}{
    >{\RaggedRight\arraybackslash}b{0.05\columnwidth}  | 
    >{\centering\arraybackslash}b{0.06\columnwidth}  | 
    X
    }
    \toprule  
    \multicolumn{3}{c}{\textbf{SDCNL}} \\
    \midrule
    \textbf{ID} & \textbf{Class} & \textbf{Text} \\
    \midrule
    \multicolumn{3}{l}{\textbf{0-10 seconds}} \\
    \midrule


    39	& DEP &	*** good qualities in therapist that i can spot early on? ***	\\
    777	& DEP &	I’m *** okay I probably won’t, but tonight *** gotta tell myself that. *** until sunrise	\\
    914	& DEP &	Another attempt ***. Hopefully I don't survive	\\
    1143	& DEP &	*** tired of being told it gets better *** never has and never *** will	\\
    1278	& DEP &	How do i know if i have depression *** i just wanna know *** i’m getting close	\\
    51	& SUI &	I’m scared *** alone *** don’t know what to do ***	\\
    599	& SUI &	Gave my note to my family *** don't know what to do now	\\
    967	& SUI &	Why do people think they can help *** on ending it all? ***	\\
    1459	& SUI &	*** sleep forever and never want to wakeup again..... ***	\\
    1494	& SUI &	Picking up a gun *** relieved	\\

    \midrule
    \multicolumn{3}{l}{\textbf{10-25 seconds}} \\
    \midrule


    562	& DEP &	Songs? *** songs/artist *** feel better/happy when you feel depressed?	\\
    786	& DEP &	*** freaking out constantly *** so lonely Even when I’m around people I’m lonely. I feel crazy. I can’t stop looking *** at my awful life, *** such poor judgment	\\
    1209	& DEP &	Anyone want to chat? I’m not in a good place *** Just wondering *** find comfort in each other.	\\
    1668	& DEP &	Any good songs *** to recommend? Any song *** which makes you feel good. Have a good day!	\\
    98	& SUI &	Suicide *** 4 ways I can go... *** overdose *** cop *** bridge *** train	\\
    251	& SUI &	Called off my work , going to end it *** Nice knowing you all I'm finally coming home *** and *** ❤️	\\
    757	& SUI &	*** want to have someone. It's all I've *** wanted *** keeps eating away at me and won't change. *** don't want to be alone anymore.	\\
    877	& SUI &	Wanna kill myself tonight. Having suicidal thoughts *** now. I wanna end it *** slit my wrists. After being on this earth *** I'm done.	\\
    889	& SUI &	*** end it all today No more suffering *** humiliation *** worrying about the future, *** such a coward i wanna die so bad	\\
    1864	& SUI &	Please someone help me I need *** an effective and painless method *** tell me I can't last longer.	\\

    \bottomrule 
\end{tabularx}
\label{tbl:audioanalysissamplessdcnl}
\end{table}
\balance

\bibliographystyle{ACM-Reference-Format}
\bibliography{sidsm}